\documentclass[10pt,twocolumn,letterpaper]{article}

\usepackage{cvpr}              

\usepackage{graphicx}
\usepackage{amsmath}
\usepackage{amssymb}
\usepackage{booktabs}

\usepackage{soul}
\usepackage{enumerate}
\usepackage{enumitem}
\usepackage{wrapfig}
\usepackage{lipsum}
\usepackage{algorithm}
\usepackage{algpseudocode}
\usepackage{caption}
\usepackage{subcaption}
\usepackage[pagebackref,breaklinks,colorlinks]{hyperref}

\usepackage[capitalize]{cleveref}
\crefname{section}{Sec.}{Secs.}
\Crefname{section}{Section}{Sections}
\Crefname{table}{Table}{Tables}
\crefname{table}{Tab.}{Tabs.}

\newcommand{\method}{FedAlign}
\newcommand{\ev}{$\lambda_{max}$}
\newcommand{\trace}{$H_{T}$}
\newcommand{\diffnorms}{$H_{N}$}
\newcommand{\cossim}{$H_{D}$}

\def\cvprPaperID{11405} 
\def\confName{CVPR}
\def\confYear{2022}

\begin{document}

\title{Local Learning Matters: Rethinking Data Heterogeneity in Federated Learning}

\author{Matias Mendieta$^{1}$, Taojiannan Yang$^{1}$, Pu Wang$^{2}$, Minwoo Lee$^{2}$, Zhengming Ding$^{3}$,  Chen Chen$^{1}$\\
$^1$Center for Research in Computer Vision, University of Central Florida, USA\\
$^2$Department of Computer Science, University of North Carolina at Charlotte, USA\\
$^3$Department of Computer Science, Tulane University, USA\\
{\tt\small \{mendieta,taoyang1122\}@knights.ucf.edu; chen.chen@crcv.ucf.edu}\\
{\tt\small \{pu.wang,minwoo.lee\}@uncc.edu; zding1@tulane.edu}
}
\maketitle

\begin{abstract}

Federated learning (FL) is a promising strategy for performing privacy-preserving, distributed learning with a network of clients (i.e., edge devices). However, the data distribution among clients is often non-IID in nature, making efficient optimization difficult. To alleviate this issue, many FL algorithms focus on mitigating the effects of data heterogeneity across clients by introducing a variety of proximal terms, some incurring considerable compute and/or memory overheads, to restrain local updates with respect to the global model. 
Instead, we consider rethinking solutions to data heterogeneity in FL with a focus on local learning generality rather than proximal restriction.
To this end, we first present a systematic study informed by second-order indicators to better understand algorithm effectiveness in FL. Interestingly, we find that standard regularization methods are surprisingly strong performers in mitigating data heterogeneity effects.
Based on our findings, we further propose a simple and effective method, \method, to overcome data heterogeneity and the pitfalls of previous methods. \method~achieves competitive accuracy with state-of-the-art FL methods across a variety of settings while minimizing computation and memory overhead. Code is available at \url{https://github.com/mmendiet/FedAlign}.

\end{abstract}
\section{Introduction}

Federated learning (FL) \cite{fedavg} enables a large number of clients to perform collaborative training of machine learning models without compromising data privacy. 
In the FL setting, participating clients are typically deployed in a variety of environments or owned by a diverse set of users. Therefore, the distribution of each client's local data can vary considerably (i.e., data heterogeneity).
This non-IID data distribution among participating devices in FL makes optimization particularly challenging. As each client trains locally on their own data, they step towards their respective local minimum. However, this local convergence point may not be well aligned with the objective of the global model (that is, the model being learned though aggregation at the central server). \textit{Therefore, the client model often drifts away from the ideal global optimization point and overfits to its local objective.} When such client drifting occurs, the performance of the central aggregated model is hindered \cite{fedml_survey, noniid_study}.

One straight-forward solution to this phenomenon is to simply limit the number of local training epochs performed between central aggregation steps. However, this severely hinders the convergence speed of the FL system, and many communication rounds are required to achieve adequate performance. The time to convergence and immense communication overhead incurred by such an approach are often not tolerable for real-world distributed systems. Therefore, effectively addressing data heterogeneity is of paramount concern in federated learning.

Many algorithmic solutions to this problem have been proposed in the literature \cite{fedprox, moon, scaffold, feddyn}. 
These strategies typically focus on mitigating the effects of data heterogeneity across clients by introducing a variety of \textit{proximal terms} to restrain local updates with respect to the global model. \textit{However, by restraining the drift, they also inherently limit the local convergence potential; less novel information is gathered per communication round.} Consequently, many current FL algorithms do not provide stable performance improvements across different non-IID settings in comparison to classic baselines \cite{noniid_study, moon}, especially on vision tasks beyond the difficulty of MNIST \cite{mnist}. Furthermore, existing methods have paid little attention to the resource constraints of the client, typically scarce for deployed FL edge devices, and in some cases incur considerable compute and/or memory overheads on the client in their effort to alleviate client drift. For example, the state-of-the-art (SOTA) method MOON performs well on federated image tasks, but to do so incurs a \textit{$\sim$3x overhead in both memory and compute} compared to the standard FedAvg baseline \cite{fedavg}.



%


\textbf{Motivation.} In the centralized training paradigm, network generalization capability has been well studied to combat overfitting. Even in standard settings where the training and test data are drawn from a similar distribution, models still overfit on the training data if no precautions are taken. This effect is further intensified when the training and test data are of different distributions. Various regularization techniques are introduced to enforce learning generality during training and preserve suitable test performance.
Similarly, overfitting to the local training data of each device in FL is detrimental to overall network performance, as the client drifting effect creates conflicting objectives among local models.
\textit{Thus, a focus on improving model generality should be of primary concern in the presence of data heterogeneity.}
Improving local learning generality during training would inherently position the objective of the clients closer to the overall global objective. However, despite its intuitive motivations, \ul{this perspective has been overlooked by the bulk of current FL literature}.




Therefore, in this paper, we propose rethinking approaches to data heterogeneity in terms of \textbf{local learning generality} rather than proximal restriction. Specifically, we carefully analyze the effectiveness of various data and structural regularization methods at reducing client drift and improving FL performance (Section \ref{empirical}). Utilizing second-order information and insights from out-of-distribution generality literature \cite{fishr, hardtovary}, we identify theoretical indicators for successful FL optimization, and evaluate across a variety of FL settings for empirical validation. 
Although some of the regularization methods perform well at mitigating client drift, \textit{significant resource overheads} are still incurred to achieve the best performance (see Section \ref{proposed-method}). 
Therefore, we propose \textbf{\method}, a distillation-based regularization method that \ul{promotes local learning generality while maintaining excellent resource efficiency}. 
Specifically, \method~focuses on regularizing the Lipschitz constants of the final block in a network with respect to its representations. 
By focusing solely on the last block, we effectively regularize the portion of the network most prone to overfitting and keep additional resource needs to a minimum. Therefore, \method~achieves state-of-the-art accuracy on multiple 
datasets across a variety of FL settings, while requiring significantly less computation and memory overhead in comparison to other state-of-the-art methods.

Our {contributions} are as follows:
\setlist{nolistsep}
\begin{itemize}[noitemsep,leftmargin=*]
    \item We 
    approach one of the most troublesome FL challenges (\ie client drift caused by data heterogeneity) from a unique angle than any other previous work. 
    We do not focus on reparameterization tricks to maintain closeness to the central model, or adjust the aggregation scheme to mitigate the effects of non-IID data distributions. \textit{Rather, we propose the rethinking of this problem from fundamental machine learning training principles.} In this way, we analyze the performance of standard regularization methods on FL and their effectiveness against data heterogeneity.
    \item Not only do we empirically analyze the performance of regularization methods in FL, we also propose to take a deeper look. Specifically, we inform our analysis with theoretical indicators of learning generality to provide insight into which methods are best and why. We find that Hessian eigenvalue/trace measurements and Hessian matching across clients to be meaningful indicators for optimal FL methods. Additionally, we perform a thorough ablation study across a variety of FL settings to understand the empirical effects of different methods.
    Our aim is to provide this valuable knowledge to the FL community to inspire new, productive research directions.
    \item Informed by our analysis and examining the pitfalls of previous methods, we propose \method, which achieves competitive state-of-the-art accuracy while maintaining memory and computational efficiency. 
\end{itemize}

\section{Related Work} \label{related}
\textbf{Federated Learning.} 
In general, federated learning algorithms aim to obtain a collective model which minimizes the training loss across all clients. This objective can be expressed as
\vspace{-0.4cm}
\begin{equation}
\min _{w} F(w)=\sum_{c=1}^{C}  \alpha_{c} F_{c}(w),
\end{equation}
where $F_{c}(w)$ is the local loss of device $c$, and $\alpha_{c}$ is an arbitrary weight parameter with $\sum_{c=1}^{C} \alpha_{c} = 1$. One of the earliest algorithms proposed in FL is Federated Averaging, or FedAvg \cite{fedavg}. This approach simply optimizes the local training loss with standard SGD training, and aggregates using a weighted average approach with $a_{c} = \frac{n_c}{n}$, where $n_{c}$ is equal to the number of training samples on client $c$, with a total of $n$ training samples partitioned across all $C$ clients.

Recent works attempt to improve over this baseline with two distinct focuses: improvements to the local training at the client, or improvements to the global aggregation process at the server. In this work, {we focus on local training and client drift}, and therefore we will first discuss methods of this nature. 
To mitigate data heterogeneity complications, a common approach is to introduce proximal terms to the local training loss. 
For instance, FedProx \cite{fedprox} 
adds the proximal term $\frac{\mu}{2}\left\|w-w^{t}\right\|^{2}$,
where $\mu$ is a hyperparameter, $w$ is the current local model weights, and $w^{t}$ is the global model weights from round $t$. The goal of this reparameterization is to minimize client drift by limiting the impact of local updates from becoming extreme.
More recently, MOON \cite{moon} proposes a similar reparameterization idea inspired by contrastive learning. Specifically, the authors form a local model constrastive loss comparing representations of three models: the global model, the current local model, and a copy of the local model from the previous round. The goals of this term are similar to that of FedProx but in feature representation space; to push the current local representation closer to the global representation. At the same time, the current local model is being pushed away from the representations of the local model copy of the previous round.
Other methods \cite{feddyn, scaffold} follow similar ideas; they aim to limit the impact of the local update or shift the update with a correction term. 

\textit{However, these approaches have two main downsides}. First, by restraining the drift, they also inherently limit the local convergence potential. With this, not as much new information is gathered per communication round. Second, many of these methods incur substantial overheads in memory and/or computation. For instance, because of its model constrastive loss, MOON \cite{moon} requires the storage of three full-size models in memory simultaneously during training, and forward passing through each of these every iteration. This requires a great deal of additional resources, which are often already scarce in FL client settings. 

Other works focus on the server side of the system, aiming to improve the aggregation algorithm. \cite{bayesian} propose a Bayesian nonparametric method for matching neurons across local models at aggregation rather than naively averaging. However, the presented framework is limited in application to fully-connected networks, and therefore \cite{fedma} extend it to CNNs and LSTMs. 
FedNova \cite{fednova} presents a normalized averaging method as an alternative to the simple FedAvg update. {As we focus on the local training, these works are orthogonal to our work.} A few approaches \cite{fedmix, 2way, xor} propose federated schemes inspired by the data augmentation method Mixup, using similar averaging techniques on the local data and sharing the augmented data with the global model or other devices. However, even though the data is augmented in some way prior to distribution, the sharing of private data from the client is less than ideal for privacy preservation. Furthermore, sharing additional data worsens the communication burden on the system, which is a principal concern in FL.

\textbf{Learning Generality.}
In traditional centralized training, the practice of regularization of various forms is common practice for improving generality. Data-level regularization, including basic data augmentations and other more advanced techniques \cite{mixup, cutmix}, are known to be quite effective. Other methods introduce a level of noise to the training process via structural modification; for instance, random or deliberate modifications to the network connectivity \cite{stochdepth, dropblock, dropout2d}. \cite{gradaug} proposes a hybrid approach that introduces self-guided gradient perturbation to the training process through the use of sub-network representations, knowledge distillation, and input transformations. As part of this work, we employ a variety of regularization methods in many FL settings and analyze their performance in comparison to state-of-the-art FL algorithms.
\section{Empirical Study} \label{empirical}
\textbf{We wish to assess the data heterogeneity challenge of FL from a simple yet unique perspective of local learning generality.} Specifically, we first study the effectiveness of standard regularization techniques as solutions to this FL challenge in comparison to state-of-the-art methods.
\subsection{Preliminaries}
We employ three FL algorithms, namely FedAvg, FedProx, and MOON. These works represent both classic baselines and current state-of-the-art, and are described in Section \ref{related}. For comparison, we employ three state-of-the-art regularization methods: Mixup \cite{mixup}, Stochastic Depth \cite{stochdepth}, and GradAug \cite{gradaug}.
Specifically, these regularization methods are applied to the local optimization within a standard FedAvg setup, and their operations are described as follows.

Mixup is a data-level augmentation technique that performs linear interpolation between two samples. Specifically, given two sample-label pairs $(x_{i}, y_{i})$ and $(x_{j}, y_{j})$, they are combined as $\tilde{x}=\beta x_{i}+(1-\beta) x_{j}$ and $\tilde{y}=\beta y_{i}+(1-\beta) y_{j}$, where $\beta \sim Beta(\gamma, \gamma)$.

Stochastic depth (StochDepth) is a structural-based method that drops layers during training, thereby creating an implicit network ensemble of different effective lengths. Specifically, the output of layer (or residual block) $\ell$ is given by
$\zeta_{\ell}=\sigma\left(\lambda \mathcal{F}_{\theta_{\ell}}\left(\zeta_{\ell-1}\right)+\mathcal{I}\left(\zeta_{\ell-1}\right)\right)$, where $\lambda$ is a Bernoulli random variable, $\mathcal{F}_{\theta_{\ell}}$ is the operation within the network with parameter $\theta$ at layer $\ell$, $\mathcal{I}$ is the identity mapping operation of residual connections, and $\sigma$ is a non-linear activation function. The keep probability is defined as $\rho = P(\lambda = 1)$, where in practice each layer has its own keep probability set with a linear decay rule $\rho_{\ell} = 1-\frac{\ell}{L}(1-\rho_{L})$, with $L$ denoting the total number of layers (or blocks) in the network.

GradAug is a recent regularization approach that combines data-level and structural techniques in a distillation-based framework. Its training loss is defined as
\begin{equation}\label{eq:gradaug}
\footnotesize
L_{G A}=L_{CE}(\mathcal{F}_{\theta}(x), y)+\mu \sum_{i=1}^{n} L_{KD}\left(\mathcal{F}_{\theta^{\omega_{i}}}\left(T^{i}(x)\right), \mathcal{F}_{\theta}(x)\right),
\end{equation}
where $\mathcal{F}_{\theta^{\omega_{i}}}$ denotes a slimmed sub-network of fractional width $\omega_{i}$, $T^{i}$ is a transformation performed on the input (e.g. resolution scaling), and $\mu$ is a balancing parameter between the cross-entropy loss $L_{CE}$ and the summed Kullback–Leibler divergence ($L_{KD}$) loss on $n$ sub-networks. The $\omega_{i}$ fractional width for each sub-network is sampled from a uniform distribution between a lower bound $\omega^{b}$ and 1.0 (full-width).
\subsection{Experimental Setup} \label{sec:setup}
To begin our analysis, we test the accuracy of several state-of-the-art FL algorithms with several regularization methods in a common FL setting. We perform experiments using CIFAR-100 \cite{cifar100}, an image recognition dataset with 50,000 training images across 100 categories, and employ ResNet56 \cite{resnet} (as implemented in FedML \cite{fedml} with PyTorch \cite{pytorch}) as the model. As common in the literature \cite{moon, feddyn, fedml}, the dataset is partitioned into $K$ unbalanced subsets using a Dirichlet distribution ($Dir(\alpha)$), with the default being $\alpha=0.5$. With this data partitioning scheme, it is possible for a client to have no samples for one or multiple classes. Therefore, many clients will only see a portion of the total class instances. This makes the setting more realistic and challenging. For all methods and experiments we use an SGD optimizer with momentum, and a fixed learning rate of 0.01. \ul{In our basic setting, training is conducted for 25 rounds, with 16 clients and 20 local epochs per round.} Any modifications to this setting in subsequent results will be stated clearly.

We compare the previously described FL algorithms and regularization methods.
FedProx, MOON, and GradAug all have a hyperparameter $\mu$ to balance their additional loss terms. We report all results with the optimal $\mu$ for all approaches, being 0.0001, 1.0, and 1.75 for FedProx, MOON, and GradAug respectively. For Mixup and Stochastic Depth, $\gamma$ and $\rho_{L}$ are set to 0.1 and 0.9 respectively. For GradAug specifically, the number of sub-networks $n=2$, $\omega^{b}=0.8$, and the applied transformation $T$ is random resolution scaling. A two-layer projection layer is added to the model for MOON and the default temperature parameter $\tau=0.5$ as specified in the original paper. Basic data augmentations (random crop, horizontal flip, and normalization) are kept consistent across all methods.

\begin{table}[h]
    \caption{Results for accuracy (\%) on CIFAR-100 and second-order metrics indicating the smoothness of the loss space (\ev, \trace) and cross-client consistency (\diffnorms, \cossim) for each method.}
    \label{cifar}
    \centering
    \small
    \renewcommand{\arraystretch}{0.8}
    \setlength\tabcolsep{3.8pt} 
    \resizebox{0.85\columnwidth}{!}{%
    \begin{tabular}{cc|cc|cc}
        \toprule
        Method & Acc. $\uparrow$ & \ev $\downarrow$ & \trace $\downarrow$ & \diffnorms $\downarrow$ & \cossim $\uparrow$\\
        \toprule
        FedAvg & 52.9 & 297 & 6240 & 11360 & 0.98\\
        FedProx & 53.0 & 270 & 6132 & 6522 & 0.98\\
        MOON & 55.3 & 252 & 5520 & 5712 & 0.97\\
        \midrule
        Mixup & 54.0 & 216 & 5468 & 15434 & 0.99\\
        StochDepth & 55.5 & 215 & 3970 & 8267 & 0.97\\
        GradAug & \textbf{57.1} & \textbf{167} & \textbf{2597} & 2924 & 0.96\\
        \midrule
    \end{tabular}}
\end{table}

\subsection{Results Comparison}
The accuracy results are shown in Table \ref{cifar}. Within the current state-of-the-art FL algorithms (upper portion of Table \ref{cifar}), MOON achieves the best accuracy. This is expected, as MOON is the most intricate of the FL methods, \ul{requiring the usage of three individual models for its contrastive learning technique.} However, when we compare with standard regularization techniques (Mixup, StochDepth and GradAug in the lower portion of Table \ref{cifar}), we see that these perform similarly or substantially better. GradAug particularly stands out, achieving an accuracy $\sim$2\% higher than MOON and $\sim$4\% higher than FedAvg and FedProx. StochDepth also achieves similar accuracy to MOON. Furthermore, these regularization methods bring the same or better performance than MOON, with less memory and/or compute requirements. \textit{We find that regularization methods appear to have an advantage in this situation; however, we wish to further investigate why this could be the case.} Next, we present our in-depth analysis based on second-order information in Section \ref{second-order-analysis}.  



\begin{figure}

\begin{center}%
\vspace{-0.1cm}
    \centering
    \begin{subfigure}[b]{0.2\textwidth}
    \centering
    \includegraphics[trim=90 50 90 90, clip, width=\textwidth]{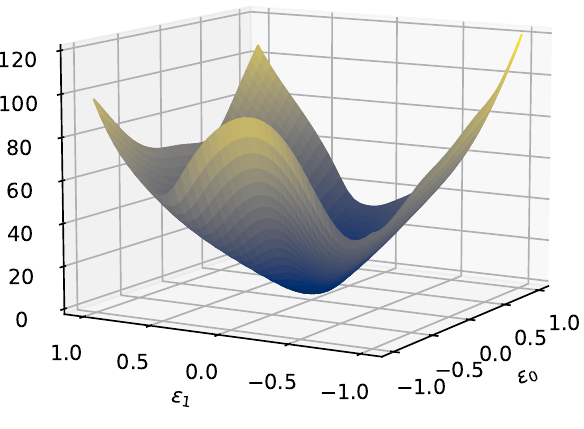}%
    \caption{FedAvg}
    \end{subfigure}
\begin{subfigure}[b]{0.23\textwidth}
    \centering
    \includegraphics[trim=120 50 20 90, clip, width=\textwidth]{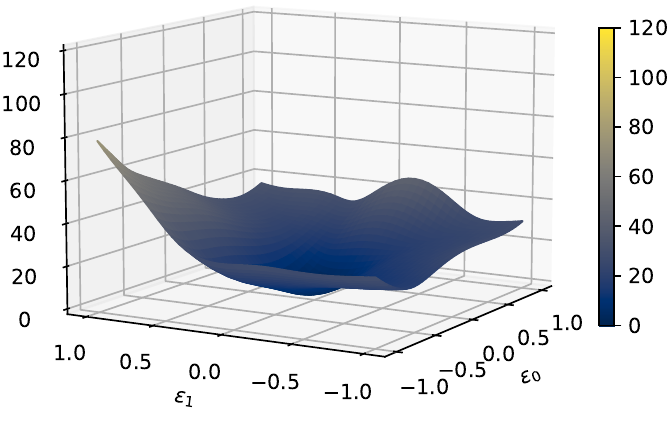}%
    \caption{GradAug}
    \end{subfigure}
    \caption{Visualization of the parametric loss landscape with Hessian eigenvectors $\epsilon_0$ and $\epsilon_1$ for each resulting global model.}%
    \label{fig:loss}%
    \vspace{-15pt}
\end{center}
\end{figure}

\subsection{Algorithm Analysis based on Second-order Information} \label{second-order-analysis}
Recent works in the Neural Architecture Search domain \cite{nas1, nas2}, as well as in network generalization \cite{sharp, power_iter, fantastic}, have noted the importance of the top Hessian eigenvalue (\ev) and Hessian trace (\trace) as a predictor of performance and indicator of network generality. Having a lower \ev~and \trace~typically yields a network that is less sensitive to small perturbations in the networks weights. 
This has the beneficial effects of smoothing the loss space during training, reaching a flatter minima, and easing convergence. These properties are particularly advantageous in federated learning, where extreme non-IID distributions and limited local data often make convergence difficult.

Motivated by these insights, we analyze the top Hessian eigenvalue and Hessian trace of the global models trained with each FL scheme to provide insight into the effectiveness of each method. 
As described in \cite{pyhessian}, the top Hessian eigenvalues can be approximated with the Power Iteration \cite{power_iter} method using a simple inner product and standard backpropagation.
Furthermore, \cite{pyhessian} also find a similar approximation for the trace utilizing the Hutchinson method \cite{hutch}. We conduct our analysis with the top Hessian eigenvalues and trace of the final averaged models using these methods.

In Table \ref{cifar}, we include the results of the Hessian analysis. First, we find that FedAvg has the highest \ev~and \trace. FedProx and MOON each result in lower values, indicating some degree of improved generalization. However, interestingly, we find that regularization methods are most effective at reducing the \ev~and \trace, with GradAug having by far the lowest in both values. We visualize the effect of this reduction in \ev~and \trace~in Fig. \ref{fig:loss}, where it can be seen that \ul{GradAug is able to smooth out the loss landscape considerably in comparison to FedAvg.}

In the separate field of out-of-distribution (O.O.D.) generalization for centralized training, second-order information is being found quite useful as a theoretical indicator. Recent works \cite{hardtovary, fishr} find that forming representations that are ``hard to vary" seem to result in better O.O.D. performance. More specifically, they show that the resulting loss landscapes across domains for the learned model should be consistent with each other. In terms of theoretical indicators, this translates to matching domain-level Hessians, as the Hessian provides an approximation of local curvature. Similarly, in federated learning, each client is essentially a separate domain. Therefore, matching Hessians in norm and direction across clients reveals additional detail and reasoning behind the effectiveness of each method. In light of these findings in O.O.D. literature, we analyze the difference in Hessian norm (\diffnorms) and the Hessian direction across clients (\cossim), where
\begin{equation}
H_{N}^{k,j} = \left(\left\|\operatorname{Diag}\left(\mathbf{H}_{k}\right)\right\|_{F}-\left\|\operatorname{Diag}\left(\mathbf{H}_{j}\right)\right\|_{F}\right)^{2} \text{and}
\end{equation}
\begin{equation}
    H_{D}^{k,j} = \frac{\operatorname{Diag}\left(\mathbf{H}_{k}\right) \odot \operatorname{Diag}\left(\mathbf{H}_{j}\right)}{\left\|\operatorname{Diag}\left(\mathbf{H}_{k}\right)\right\|_{F} \cdot \left\|\operatorname{Diag}\left(\mathbf{H}_{j}\right)\right\|_{F}}.
\end{equation}
Here, $\odot$ is the dot product, $\mathbf{H}_{k}$ and $\mathbf{H}_{j}$ are the Hessian matrices of clients $k$ and $j$, and $\left\| \cdot \right\|_{F}$ is the Frobenius norm. $H_{N}^{k,j}$ and $H_{D}^{k,j}$ are averaged across all pairs of clients and reported as simply \diffnorms~and \cossim~in Table \ref{cifar}. For these Hessian matching criteria, a lower \diffnorms~(less difference) and a higher \cossim~(essentially the cosine similarity) are desired.

As seen on the right side of Table \ref{cifar}, \cossim~is fairly consistent across all methods. In terms of \ev, \trace, and \cossim, most methods seem to correlate decently well between these values and performance. However, there are a few cases which require more information. First, Mixup has a similar \trace~value as MOON, but lower accuracy. \diffnorms~provides another detail; the Hessian norms of Mixup are not nearly as similar across clients as those of MOON. Between MOON  and StochDepth, we see that MOON has both a higher \ev~and \trace, but StochDepth has a higher \diffnorms. In the end, MOON and StochDepth result in similar performance, with perhaps a slight edge towards the latter.

\textbf{Key Insight.} It appears that both the eigenvalue/trace analysis and Hessian matching criteria can serve as a guiding indicator for optimal FL methods. Particularly, they provide insight into the facilitation of convergence and aggregation thorough landscape smoothness and consistency. To understand how these differences will play out empirically, we conduct a variety of ablations in Section \ref{exp-ablation}.


\subsection{Ablation Study under Various FL Settings} \label{exp-ablation}
\noindent \textbf{Data Heterogeneity.}
Federated systems can be deployed with many different setups and diverse environments. We conduct further analysis across a variety of FL settings to ensure the generality of our findings. First, we examine the effect of varying the degree of heterogeneity in the client data distributions. The results are shown in Table \ref{tab:data-ablation}. We report the mean accuracy $\pm$ the standard deviation across three runs. All other settings are maintained from Section \ref{sec:setup}; only the data distribution $Dir(\alpha)$ is varied. \textit{A lower $\alpha$ value indicates a more heterogeneous distribution.}
\begin{table}[h]
    \caption{Ablation results for varying degrees of data heterogeneity.}
    \label{tab:data-ablation}
    \centering
    \setlength\tabcolsep{3.3pt} 
    \renewcommand{\arraystretch}{0.8}
    \small
    \begin{tabular}{ccccc}
        \toprule
        Method & $\alpha=0.1$ & $\alpha=0.5$ & $\alpha=2.5$ & homog\\
        \toprule
        FedAvg & 45.0$\pm$0.2 & 52.9$\pm$0.1 & 54.4$\pm$0.2 & 54.9$\pm$0.4\\
        FedProx & 45.2$\pm$0.3 & 53.1$\pm$0.3 & 54.5$\pm$0.3 & 54.8$\pm$0.5\\
        MOON & 46.5$\pm$0.5 & 55.0$\pm$0.5 & 56.3$\pm$0.6 & 56.3$\pm$0.5\\
        \midrule
        Mixup & 44.3$\pm$0.1 & 54.0$\pm$0.1 & 55.5$\pm$0.4 & 56.7$\pm$0.4\\
        StochDepth & 48.2$\pm$0.3 & 55.5$\pm$0.2 & 57.6$\pm$0.2 & 58.1$\pm$0.6\\
        GradAug & \textbf{48.6$\pm$0.4} & \textbf{57.0$\pm$0.1} & \textbf{59.6$\pm$0.2} & \textbf{60.5$\pm$0.2}\\
        \midrule
    \end{tabular}
    \vspace{-10pt}
\end{table}

As the degree of data heterogeneity decreases, the effect of client drift should become less significant. Therefore, we expect that the accuracy for each method will increase, with peak performance in the homogeneous setting. All regularization methods, as well as FedAvg, perform as expected, and find consistent improvement across the degrees of data distribution. However, we see that the accuracy improvement of FedProx and MOON slows as the data approaches homogeneity, with accuracy in the purely homogeneous setting (``homog" in Table \ref{tab:data-ablation}) remaining quite low. In their attempt to mitigate client drift and keep local updates close to the global model, it appears that they also hinder their ability to fully learn on minorly heterogeneous or even homogeneous data. This is not ideal for deployable FL systems, as the degree of heterogeneity is not known ahead of time. Moreover, even in the most heterogeneous cases, the structural regularization methods perform better than the standard FL algorithms. For instance, StochDepth achieves a $\sim$1.7\% improvement over MOON at $\alpha$ = $0.1$, while also having improvement in more homogeneous situations.
In all settings, GradAug performs the best.

\begin{table}
    \caption{Ablation results for number of local training epochs.}
    \label{tab:epochs}
    \centering
    \setlength\tabcolsep{5.0pt} 
    \renewcommand{\arraystretch}{0.8}
    \small
    \begin{tabular}{cccc}
        \toprule
        Method & $E=10$ & $E=20$ & $E=30$\\
        \toprule
        FedAvg & 50.6$\pm$0.1 & 52.9$\pm$0.1 & 53.2$\pm$0.3\\
        FedProx & 50.7$\pm$0.5 & 53.1$\pm$0.3 & 52.8$\pm$0.1\\
        MOON & 50.7$\pm$0.4 & 55.0$\pm$0.5 & 55.2$\pm$0.4\\
        \midrule
        Mixup & 50.5$\pm$0.4 & 54.0$\pm$0.1 & 54.4$\pm$0.3\\
        StochDepth & 50.9$\pm$0.6 & 55.5$\pm$0.2 & 56.4$\pm$0.3\\
        GradAug & \textbf{53.5$\pm$0.3} & \textbf{57.0$\pm$0.1} & \textbf{57.7$\pm$0.3}\\
        \midrule
    \end{tabular}
    \vspace{-15pt}
\end{table}


\noindent \textbf{Number of Local Training Epochs.}
The main purpose for adequately handling data heterogeneity is to allow for more productive training on the client each round, therefore reducing the time to convergence and required communication cost. Therefore, to examine the training productivity of each method, we examine their accuracy with various allotted local training epochs per round ($E$) in Table \ref{tab:epochs}. 


Ideally methods should continue to improve in accuracy with more allotted local training epochs. In Table \ref{tab:epochs}, we see that all methods steadily improve from 10 epochs per round to 20. However, from 20 to 30, the trends vary considerably. As a baseline, FedAvg slightly improves by $\sim$0.3\%. Surprisingly, FedProx and MOON stay relatively stagnant from 20 to 30 epochs. Meanwhile, the standard (particularly structural) regularization methods continue to increase in accuracy. Therefore, these methods illustrate the ability to maintain productive training, even across a wide range of allotted local epochs.

\noindent \textbf{Number of Clients.}
In real-world FL settings, the number of participating clients can vary widely. Moreover, only a portion of clients are potentially sampled per round, whether for connectivity reasons or other capacity restrictions of the central system. Therefore, it is crucial that an FL method can converge under such conditions. We study the affect of client number and client sampling in Table \ref{tab:clients-num}. $C=64\times0.25$ indicates that there are 64 total clients in the system, but only a fraction (0.25) are sampled each round. The rest of the presented results in Table \ref{tab:clients-num} sample all $K$ clients each round. $C=64\times0.25$ (100) is run for 100 rounds, and all other settings for the default 25 rounds. 

\begin{table}[t]
    \caption{Ablation results for varying number of clients $C$ in synchronous and client sampling cases.}
    \label{tab:clients-num}
    \centering
    \small
    \renewcommand{\arraystretch}{0.8}
     \resizebox{\linewidth}{!}{
    \begin{tabular}{cccccc}
        \toprule
        Method & $C=16$ & $C=32$ & $C=64$ & $C=64\times0.25$ & $C=64\times0.25$ (100)\\
        \toprule
        FedAvg & 52.9$\pm$0.1 & 44.5$\pm$0.3 & 34.6$\pm$0.2 & 32.7$\pm$0.5 & 46.5$\pm$0.6\\
        FedProx & 53.1$\pm$0.3 & 44.5$\pm$0.6 & 34.8$\pm$0.2 & 32.5$\pm$0.4 & 46.2$\pm$0.1\\
        MOON & 55.0$\pm$0.5 & 45.8$\pm$0.3 & 35.2$\pm$0.8 & 34.2$\pm$0.2 & 49.5$\pm$0.7\\
        \midrule
        Mixup & 54.0$\pm$0.1 & 46.0$\pm$0.1 & 36.0$\pm$0.2 & 33.6$\pm$0.6 & 49.1$\pm$0.2\\
        StochDepth & 55.5$\pm$0.2 & 47.5$\pm$0.2 & 35.5$\pm$0.6 & 34.6$\pm$0.1 & 51.4$\pm$0.1\\
        GradAug & \textbf{57.0$\pm$0.1} & \textbf{50.4$\pm$0.1} & \textbf{40.2$\pm$0.1} & \textbf{38.1$\pm$0.3} & \textbf{53.3$\pm$0.5}\\
        \midrule
    \end{tabular}
    }
\end{table}


The trends of most methods are similar with increasing clients. However, FedProx struggles to keep up with the FedAvg baseline, especially in the client sampling cases. These scenarios are particularly important; when a small percentage of clients are sampled, only a portion of the dataset is effectively trained on each round. Therefore, learning efficiency becomes paramount for maintaining suitable convergence.
The standard regularization methods maintain better accuracy than FedAvg in all settings, often by a significant margin, and even in the client sampling scenario. 
Overall, GradAug performs the best in all cases. \textit{Therefore, even though these regularization methods were not designed for the FL setting and partial client sampling, they still perform on par with or improve over current state-of-the-art FL algorithms.}

\section{Proposed Method -- \method} \label{proposed-method}



Overall, we find that GradAug is particularly effective in the FL setting, having the highest accuracy in all tested scenarios along with the lowest \ev, \trace, and \diffnorms. 
However, while this method is quite memory efficient in comparison to many FL methods (only requires a single stored model during training), it does incur a substantial increase in training time and local computation over the FedAvg baseline. This is because GradAug requires multiple forward passes through slimmed sub-networks for the distillation loss. It is possible to reduce the computation burden to some extent by using a smaller number of sub-networks during the knowledge distillation process, as seen in Table \ref{tab:walltime}. Here, the $\mu$ in GradAug is adjusted to 2.0, 1.5, and 1.25 for $n$ = 1, 3, and 4, respectively. Nonetheless, a considerable gap still remains between GradAug and vanilla FedAvg in local compute requirements and subsequent wall-clock time.
\textbf{Therefore, the question is, can we devise a method which provides similar effect and performance as GradAug in FL, but with substantially less computational overhead?} This is particularly important in the FL setting, where clients are typically deployed devices with minimal memory and computational resources. 


\begin{table}[h]
    \caption{Analysis of local compute, stored parameters, and wall-clock time. FLOPs are calculated for the compute needs for the forward pass of the training process. Parameters include the total number of stored parameters needed for each method during training. Wall-clock time is measured as a per-round average on CIFAR-100 with $C$=16 and $E$=20 across 4 RTX-2080Ti GPUs.}
    \label{tab:walltime}
    \centering
    \small
    \setlength\tabcolsep{2.6pt} 
    \renewcommand{\arraystretch}{0.8}
    \resizebox{0.9\columnwidth}{!}{
    \begin{tabular}{ccccc}
        \toprule
        Method & Acc (\%) $\uparrow$ & MFLOPs $\downarrow$ & Param (M) $\downarrow$ & Time (s)\\
        \toprule
        FedAvg & 52.9$\pm$0.1 & 87.3 & \textbf{0.61} & 137.2\\
        FedProx & 53.1$\pm$0.3 & 87.3 & 1.21 & 161.9\\
        MOON & 55.0$\pm$0.5 & 262.2 & 2.21 & 414.2\\
        \midrule
        Mixup & 54.0$\pm$0.1 &  87.3 & \textbf{0.61} & 137.8\\
        StochDepth & 55.5$\pm$0.2 & \textbf{82.4} & \textbf{0.61} & \textbf{136.7}\\
        \midrule
        GradAug ($n=1$) & 56.7$\pm$0.3 & 133.9 & \textbf{0.61} & 229.2\\
        GradAug ($n=2$) & \textbf{57.0$\pm$0.1} & 170.7 & \textbf{0.61} & 323.9\\
        GradAug ($n=3$) & 56.8$\pm$0.3 & 217.4 & \textbf{0.61} & 417.7\\
        GradAug ($n=4$) & 56.9$\pm$0.3 & 264.1 & \textbf{0.61} & 514.4\\
        \midrule
        \textbf{\method} & 56.9$\pm$0.5 & 89.1 & \textbf{0.61} & 166.2\\
        \midrule
    \end{tabular}}
\end{table}

To do so, we first take note of the following insights gathered during our analysis: 1) Second-order information is insightful for understanding the learning generality of neural networks. 
Particularly, we find that flatness and consistency in this realm are desirable traits. 2) In practice, we find that structural regularization, and especially distillation-based like GradAug, is quite effective. Furthermore, the weight sharing mechanisms of such approaches are memory efficient compared to other methods that rely on global model or previous model storage. Therefore, we combine these insights into a novel algorithm to optimize for performance and resource needs in FL. 
\begin{figure}
\centering
    \includegraphics[width=0.7\linewidth]{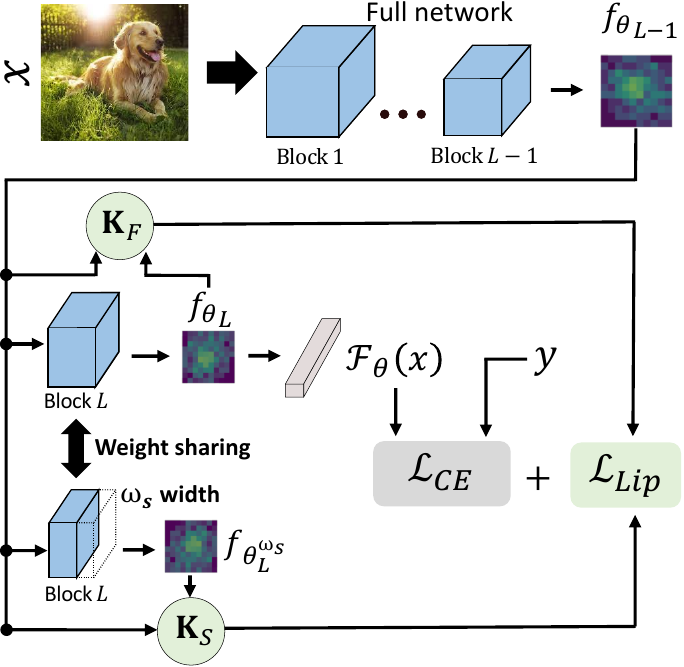}
  \vspace{-5pt}
    \caption{The proposed \method~for local client training in FL. Features $f_{\theta_{L-1}}$ are run through Block $L$ as normal. The only additional inference in \method~is through Block $L$ at a reduced width (i.e. sub-block), reusing features $f_{\theta_{L-1}}$ as input. The channels throughout the layers in the sub-block are a $\omega_{S}$ fraction of the original number. This is accomplished via temporary uniform pruning of Block $L$.}
\label{fig:method}
\vspace{-15pt}
\end{figure}

We propose \textbf{\method}, a distillation-based regularization method that aligns the Lipschitz constants (i.e. top Hessian eigenvalues) of the most critical network components through the use of slimmed sub-blocks. Fig. \ref{fig:method} shows an overview of \method, whose design
is based on \ul{two key principles}. First, motivated by the insights of Section \ref{second-order-analysis}, we internally regularize the Liptschitz constants of network blocks to
\textit{promote smooth optimization and consistency within the model.}
Recent work 
\cite{london} presents a quick approximation of the Lipschitz constants 
for neural network layers in a differentiable manner. This enables the use of second-order information in the distillation process, traditionally between a fully trained teacher and a learning student. We adapt this technique for distillation-based regularization with a \textit{single untrained network} in place of the traditional logit-based loss.
Second, in order to reduce computation in a purposeful manner, we take note of certain network properties. Particularly, it has been shown that the final layers of a neural network are most prone to overfit to the client distribution \cite{calibration}. 
Therefore, we design \method~with a focus on these critical points in the network.
\textit{The question we raise is, when aiming to concentrate our regularization efforts on the final layers, why should we run all sub-networks for distillation from start to finish?} 
Instead, we propose to reuse the intermediate features of the full network as input to just the final block at a reduced width, and therefore significantly reduce computation.
In this way, we harness the benefits of distillation-based regularization in performance and memory footprint, while effectively mitigating computational overhead.

Combining these two key principles, we form the \method~local objective as
\begin{equation}\label{eq:method}
\mathcal{L}_{FA}=\mathcal{L}_{CE}(\mathcal{F}_{\theta}(x), y)+\mu \mathcal{L}_{Lip}\left(\mathbf{K}_{S}, \mathbf{K}_{F}\right),
\end{equation}
where $\mu$ is a balancing constant, $\mathcal{L}_{CE}$ is the cross-entropy loss, and $\mathcal{L}_{Lip}$ is the mean squared error between the approximated Liptschitz constant vectors $\mathbf{K}_{S}$ and $\mathbf{K}_{F}$ for the reduced width (i.e. sub-block) and full width block $L$, respectively. Specifically, the Lipschitz approximations are calculated via the spectral norm of a transmitting matrix using feature maps as in \cite{london}, which bypasses the need for singular value decomposition. Therefore, we use the intermediate features for these transmitting matrices $\mathbf{X}_{F}$ and $\mathbf{X}_{S}$, where
\begin{equation}\label{eq:xf}
\mathbf{X}_{F} = \left(f_{\theta_{L-1}}\right)^{\top} f_{\theta_{L}}, \text{and}
\end{equation}
\begin{equation}\label{eq:xs}
\mathbf{X}_{S} = \left(f_{\theta_{L-1}}\right)^{\top} f_{\theta_{L}^{\omega_{S}}}.
\end{equation}
$f_{\theta_{L}}$ and $f_{\theta_{L-1}}$ are the feature maps outputted by the last and prior-to-last blocks of the full network $\mathcal{F}_{\theta}(x)$; $f_{\theta_{L}^{\omega_{S}}}$ is the output feature map of the final block $L$ at reduced width $\omega_{S}$ (see Fig. \ref{fig:method}). Finally, the spectral norm ($SN$) of $\mathbf{X}_{F}$ and $\mathbf{X}_{S}$ are approximated using the Power Iteration method \cite{power_iter}, and therefore $\mathbf{K}_{F} =\left\|\mathbf{X}_{F}\right\|_{SN}$ and $\mathbf{K}_{S} =\left\|\mathbf{X}_{S}\right\|_{SN}$. A pseudocode implementation of FedAlign in presented in Alg. \ref{alg:fedalign}. Looking back to Eq. \ref{eq:method}, one could view $\mathcal{L}_{Lip}$ as a correction term; however, there is a key distinction between this form of regularization and that of traditional FL algorithms. \textit{Our correction term promotes the local client models to learn well-generalized representations based on their own data, instead of forcing the local models to be close to the global model.}

\begin{algorithm}[t]
\scriptsize
\caption{FedAlign}\label{alg:fedalign}

    \begin{algorithmic}
    \vspace{0.5mm}
    \hspace{-8mm}\colorbox[gray]{0.95}{
    \begin{minipage}{0.94\columnwidth}
        \State \textbf{\textsc{Server Operations}}
        \State \textbf{Inputs:} Round number $R$, Set of clients $S$
        \State \textbf{Output:} Final global model weights $\theta^{R}_{global}$
        \vspace{1mm}
        \State Initialize model weights $\theta^{0}_{global}$
        \For{$r = 0,1,\dots, R-1$}
            \State Sample available clients $C$ from $S$
            \For {client $c \in C$ \textbf{in parallel}}
                \State $\theta^{r}_{c} \gets \textsc{ClientOperations}(\theta^{r}_{global})$
            \EndFor
            \vspace{1mm}
            \State $\theta^{r+1}_{global} \gets \sum_{c=1}^{C}  \frac{n_c}{n}\theta^{r}_{c}$
            \vspace{1mm}
        \EndFor
        \end{minipage}}
        \vspace{1mm}
        \State 
        \hspace{-3mm}\colorbox[rgb]{0.95, 0.98, 1.0}{
        \begin{minipage}{0.94\columnwidth}
        \State \textbf{\textsc{\textcolor{blue}{Client Operations}}}
        \State \textbf{Input:} Model weights $\theta_{global}$
        \State \textbf{Output:} Updated local model weights $\theta$
        \vspace{1mm}
        \State Load received weights $\theta_{global}$ to local model $\mathcal{F}_{\theta}$
        \For{$epoch~e = 0, 1 \dots, E-1$}
        \For{batch $\{x,y\}~\in~D$} \Comment{Local dataset $D$}
            \State  $f_{\theta_{L-1}}, f_{\theta_{L}}, pred = \mathcal{F}_{\theta}(x)$
            \State $f_{\theta_{L}^{\omega_{S}}} = \mathcal{F}_{\theta_{L}^{\omega_{S}}}(f_{\theta_{L-1}})$
            \State $\mathbf{X}_{S}, \mathbf{X}_{F} = TM(f_{\theta_{L}^{\omega_{S}}}, f_{\theta_{L-1}}, f_{\theta_{L}})$ \Comment{Eqs. \ref{eq:xf}, \ref{eq:xs}}
            \State $\mathbf{K}_{S}, \mathbf{K}_{F} = \left\|\mathbf{X}_{S}\right\|_{SN}, \left\|\mathbf{X}_{F}\right\|_{SN}$
            \State $\mathcal{L}_{FA}=\mathcal{L}_{CE}(pred, y)+\mu \mathcal{L}_{Lip}\left(\mathbf{K}_{S}, \mathbf{K}_{F}\right)$
            \State $\theta \gets update(\theta, \mathcal{L}_{FA})$ \Comment{Gradient descent}
        \EndFor
        \EndFor
        \State Send updated local model weights $\theta$ to server
        \end{minipage}}
    \end{algorithmic}
\end{algorithm}


\begin{table*}[!htp]
    \caption{\method~ablation results on CIFAR-100.}
    \label{tab:method}
    \centering
    \resizebox{0.85\textwidth}{!}{%
    \begin{tabular}{c|ccc|cc|cccc}
        \toprule
        Method & $\alpha=0.1$ & $\alpha=2.5$ & homog  & $E=10$ & $E=30$ & $C=32$ & $C=64$ & $C=64\times0.25$ & $C=64\times0.25$ (100)\\
        \toprule
        \method & 48.7$\pm$0.2 & 57.6$\pm$0.6 & 58.2$\pm$0.1 & 51.2$\pm$0.3 & 57.9$\pm$0.6 & 47.8$\pm$0.3 & 36.5$\pm$0.1 & 34.9$\pm$0.6 & 50.9$\pm$0.5\\
        \midrule
    \end{tabular}
    }
\end{table*}

\begin{table*}[!htp]
\footnotesize
    \caption{CIFAR-10 and ImageNet-200 results for all methods.}
    \label{tab:img200}
    \centering
    \renewcommand{\arraystretch}{0.8}
    \resizebox{0.85\textwidth}{!}{%
    \begin{tabular}{ccccc|cccc}
        \toprule
               & \multicolumn{4}{c|}{CIFAR-10} & \multicolumn{4}{c}{ImageNet-200}\\
        Method & $C=16$ & $C=64\times0.25$ (100) & MFLOPs $\downarrow$ & Param (M) $\downarrow$ & $C=16$ & $C=32\times0.125$ (50) & GFLOPs $\downarrow$ & Param (M) $\downarrow$\\
        \toprule
        FedAvg & 81.9$\pm$0.6 & 78.9$\pm$0.3 & 87.3 & 0.61 & 60.7$\pm$0.4 & 52.7$\pm$0.2 & 18.1 & 11.22\\
        FedProx & 81.9$\pm$0.2 & 78.9$\pm$0.7 & 87.3 & 1.21 & 61.0$\pm$0.4 & 52.5$\pm$0.3 & 18.1 & 22.42\\
        MOON & 82.9$\pm$0.4 & 79.4$\pm$0.5 & 262.2 & 2.21 & 61.1$\pm$0.2 & 54.3$\pm$0.2 & 54.4 & 19.96\\
        \midrule
        Mixup & 80.3$\pm$0.4 & 80.5$\pm$0.5 & 87.3 & 0.61 & 61.0$\pm$0.3 & 52.3$\pm$0.3 & 18.1 & 11.22\\
        StochDepth & 82.2$\pm$0.2 & 80.8$\pm$0.7 & 82.4 & 0.61 & 60.5$\pm$0.2 & 52.9$\pm$0.2 & 17.3 & 11.22\\
        GradAug ($n=2$) & 84.6$\pm$0.6 & 83.8$\pm$0.3 & 170.7 & 0.61 & 63.5$\pm$0.4 & 55.6$\pm$0.1 & 34.4 & 11.22\\
        GradAug ($n=1$) & 84.0$\pm$0.2 & 82.3$\pm$0.5 & 133.9 & 0.61 & 62.8$\pm$0.3 & 54.4$\pm$0.4 & 25.3 & 11.22\\
        \midrule
        \textbf{\method} & 82.3$\pm$0.3 & 82.3$\pm$0.3 & 89.1 & 0.61 & 62.0$\pm$0.1 & 55.1$\pm$0.5 & 19.3 & 11.22\\
        \bottomrule
    \end{tabular}}
\end{table*}
As seen in Table \ref{tab:walltime}, \method~achieves state-of-the-art accuracy in a resource-efficient manner. 
With just a 1.02x difference in FLOPs, \method~realizes a significant $\sim$4.0\% accuracy improvement over the FedAvg baseline.
For the FL algorithms FedProx and MOON, they not only have much lower accuracy than \method, but also require substantially more compute and/or memory. Particularly, \method~achieves a $\sim$1.9\% accuracy improvement over MOON, while reducing the local compute overhead by over 65\% and the memory requirements by over 70\%. Furthermore, \method~realizes a critical $\sim$47\% and $\sim$33\% reduction in compute needs compared to GradAug with ($n=2$) and ($n=1$), without sacrificing accuracy. 




\subsection{\method~Experiments} \label{method_exp}
We further verify the effectiveness of our method across various settings and datasets. In Table \ref{tab:method}, we examine the performance of \method~with the same ablations as in Section \ref{exp-ablation}, where \method~exhibits strong performance in many settings. We also investigate \method~and all other methods across two additional datasets: CIFAR-10 and ImageNet-200. For ImageNet-200, we randomly sample 200 classes from the classic ImageNet-1k \cite{IN1k} dataset. We employ ResNet56 and ResNet18 \cite{resnet} as our models on CIFAR-10 and ImageNet-200, respectively. For \method, $\omega_{S}=0.25$ and $\mu=0.45$ in all results. 
Hyperparameters for all other methods are those described in Section \ref{sec:setup} (with $\mu=2.0$ for GradAug ($n=1$) as in Table \ref{tab:walltime}). For additional analyses, please refer to the supplementary material.

For CIFAR-10, we ran a 16 client synchronous and 64 client case with sampling in Table \ref{tab:img200}. We note similar trends to CIFAR-100; regularization methods perform well, particularly in the more realistic client sampling case. On ImageNet-200, we also ran synchronous and sampling settings. Here, both GradAug and \method~maintain higher performance than other methods. \method~provides competitive accuracy with GradAug ($n=1$) and even ($n=2$) in the sampling case, while reducing computational needs by a significant margin. Interestingly, StochDepth does not perform as well in the ImageNet-200 cases. As mentioned in the original paper \cite{stochdepth}, Stochastic Depth performs better with deeper networks. However, with ResNet18, the overall depth of the network is reduced compared to that in the CIFAR cases. Therefore, as most deployable networks favor width over depth, regularizing with respect to the width of a network is more applicable to the FL setting. \textit{This highlights an additional benefit of \method, which operates using width reduction in the final block and maintains relatively high accuracy despite low resource needs.}

\section{Conclusion and Discussion}
In this work, we study the data heterogeneity challenge of FL from a simple yet unique perspective of local learning generality. 
To this end, we present a thorough study of various methods in FL settings, and further propose \method, which achieves competitive SOTA accuracy with excellent resource efficiency. 
One limitation of our study is that we only focused on image tasks and models for the experiments. Natural language processing applications of FL are also a common setting, and therefore could be explored in future work. Nonetheless, we note that \method~can easily be applied to language applications, as it operates in the feature space and does not have a fundamental reliance on the input type. On the other hand, GradAug is primarily designed for vision data, employing a random transformation and applying it to the input of sub-networks.

While no one presented regularization method is perfect in all respects, we emphasis that local learning is extremely important in federated settings. Furthermore, methods that particularly focus on promoting learning generality inherently improve global FL aggregation and optimization to a surprising degree. By introducing methods like GradAug in FL, we propose a rethinking of federated optimization and how to tackle its challenges. 
As a step further in this direction, FedAlign provides strong improvement over classic baselines and state-of-the-art FL methods while addressing the local computational restraints of an FL system.\\
\textbf{Acknowledgement:} This work is supported by the NSF/Intel Partnership on MLWiNS under Grant No. 2003198.

{\small
\bibliographystyle{ieee_fullname}
\bibliography{egbib}
}

\clearpage
\appendix
\section*{Supplementary Material}

The supplementary material is organized into the following sections:

\begin{itemize}
    \item Section~\ref{sec:comm}: Analysis of both communication and compute efficiency for all explored methods.
    \item Section~\ref{sec:order}: Second-order analysis of \method.
    \item Section~\ref{sec:hyper}: Hyperparameter ablations for \method.
    \item Section~\ref{sec:hyper}: Details and visualization for the non-IID data partitioning scheme.
    \item Section~\ref{sec:details}: Implementation details for transmitting matrices and \method~training.

\end{itemize}

\section{Communication and Compute Efficiency} \label{sec:comm}
Communication cost is another critical factor in FL systems, as participating client devices are often on slow or congested networks \cite{fedavg}.
Therefore, total efficiency in FL systems includes both the ability to reduce the local computational burden, as well as the communication overhead. We evaluate all methods with such measures in Table \ref{tab:comm}. We maintain the CIFAR-100 setting described in Section 3.2 of the main paper, except that we do not limit the number of rounds, but rather allow all methods to reach a common accuracy of 60\%. This allows us to analyze the total costs of each method consistently. \method~proves to be the most efficient in all respects, achieving the target accuracy in less number of rounds, less communication cost, and less local computation.
\begin{table}[h]
    \caption{Number of rounds (Rounds), local compute (MFLOPs), and communication cost (Comm Cost) required by each method to achieve 60\% accuracy on CIFAR-100. Local computation is computed as the sum total over all nodes and samples for all completed rounds. Communication costs are calculated by the number of parameters of the model transferred as 32 bit weights with all completed rounds.}
    \label{tab:comm}
    \centering
    \setlength\tabcolsep{3.3pt} 
    \renewcommand{\arraystretch}{0.8}
    \small
    \begin{tabular}{cccc}
        \toprule
        Method & Rounds & MFLOPs & Comm Cost (Gb)\\
        \toprule
        FedAvg & 84 & 7332 & 26.23\\
        FedProx & 78 & 6809 & 24.36\\
        MOON & 55 & 14421 & 17.18\\
        \midrule
        Mixup & 71 & 6198 & 22.17\\
        StochDepth & 46 & 3790 & 14.37\\
        GradAug ($n=2$) & 41 & 6999 & 12.81\\
        GradAug ($n=1$) & 44 & 5892 & 13.74\\
        \midrule
        \textbf{FedAlign} & \textbf{37} & \textbf{3297} & \textbf{11.56}\\
        \bottomrule
    \end{tabular}
\end{table}

\section{Second-order Analysis} \label{sec:order}
In Table \ref{tab:hess}, we show the second-order analysis results for \method~along with the other methods. The Liptzshitz-focused distillation loss of \method~ effectively reduces the Lipshitz constant \ev~considerably as intended (\method~results in the lowest \ev~across all methods), and therefore helps provide stronger generalization and performance. 
However, one limitation to \method~is that it does not directly translate to a strong reduction in \diffnorms.
Therefore, a promising direction for future work could extend \method~to also consider this aspect. 


\begin{figure*}

\begin{center}%
    \centering
    \subfloat[\centering CIFAR-100, $\alpha=0.5$]{{\includegraphics[width=0.9\columnwidth]{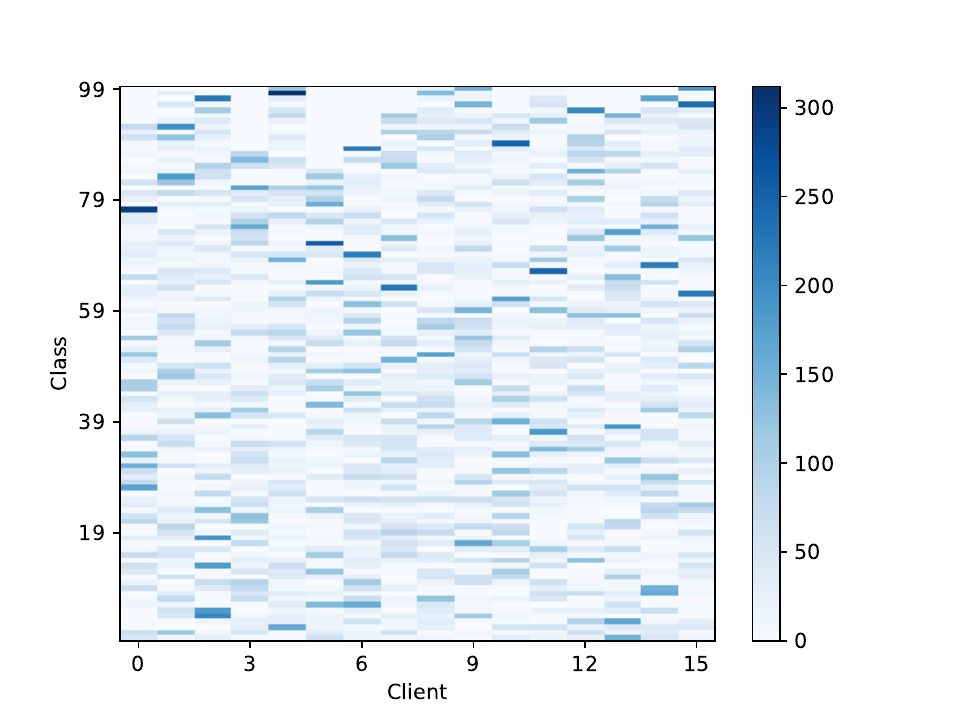} }}%
    \subfloat[\centering CIFAR-100, $\alpha=0.1$]{{\includegraphics[width=0.9\columnwidth]{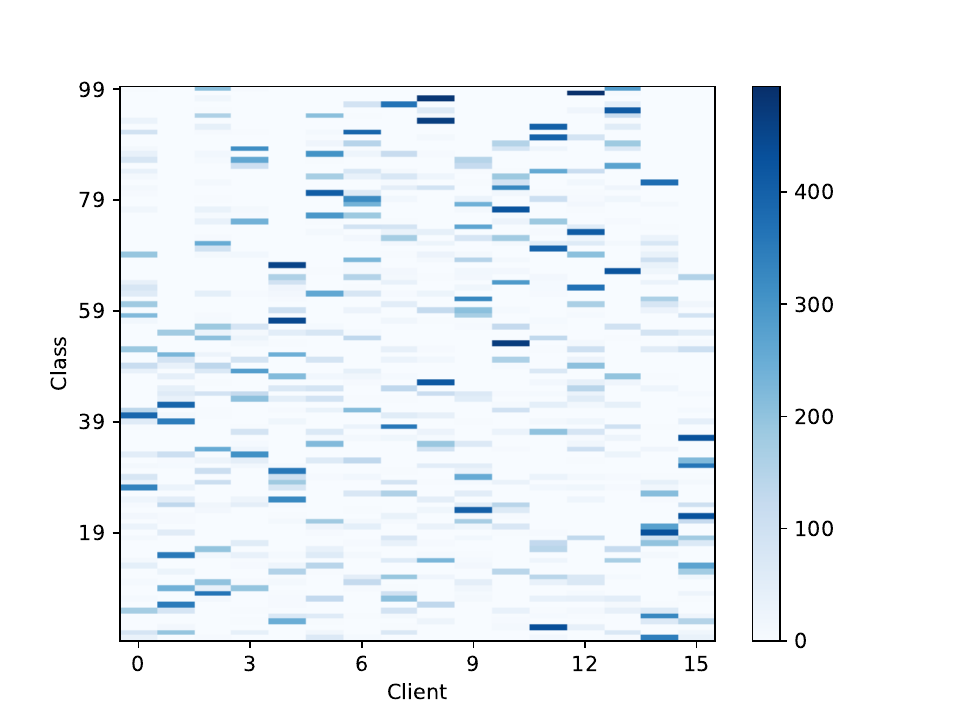} }}\\%
    \subfloat[\centering CIFAR-10, $\alpha=0.5$]{{\includegraphics[width=0.9\columnwidth]{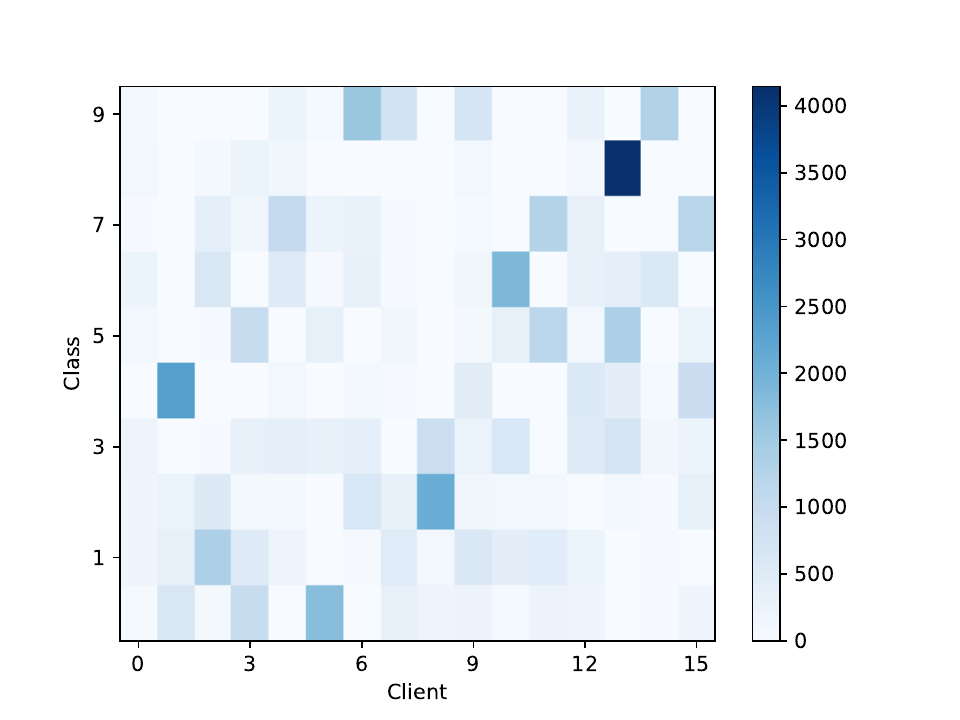} }}%
    \subfloat[\centering ImageNet-200, $\alpha=0.5$]{{\includegraphics[width=0.9\columnwidth]{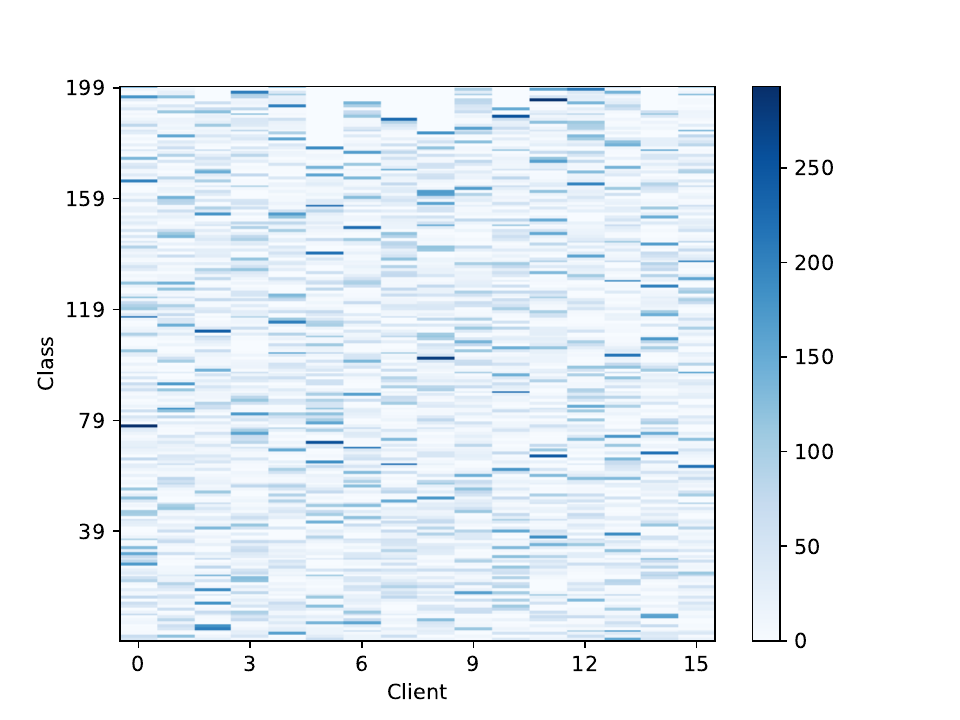} }}%

    \caption{Data distribution visualization for $Dir(\alpha)$ and $C=16$ across multiple datasets. Each column shows the number of samples per class allocated to a client.}%
    \label{fig:data}%
\end{center}
\end{figure*}

\section{Hyperparameter Ablations of FedAlign}
\label{sec:hyper}
The default hyperparameter setting used throughout the paper is $\omega_{S}=0.25$ and $\mu=0.45$. The performance of \method~with various hyperparameters on the CIFAR-100 basic setting (described in Section 3.2 of the main paper) is shown in Table \ref{tab:hyper}. We vary $\mu$ and $\omega_{S}$ independently, meaning $\omega_{S}=0.25$ for the $\mu$ ablations, and $\mu=0.45$ when varying $\omega_{S}$.
Table \ref{tab:hyper} shows that \method~is more sensitive to $\omega_{s}$ than $\mu$; nonetheless, we found $\omega=0.25$ to be a versatile choice in practice.
Furthermore, hyperparameters only need to be decided once, as they transfer well across a variety of other datasets and FL settings as shown in Section 4.1 of the main paper.
\begin{table}[h]
    \caption{Results for accuracy (\%) on CIFAR-100 and second-order metrics indicating the smoothness of the loss space (\ev, \trace) and cross-client consistency (\diffnorms, \cossim) for each method.}
    \label{tab:hess}
    \centering
    \small
    \renewcommand{\arraystretch}{0.8}
    \setlength\tabcolsep{3.8pt} 
    \resizebox{0.85\columnwidth}{!}{%
    \begin{tabular}{cc|cc|cc}
        \toprule
        Method & Acc. $\uparrow$ & \ev $\downarrow$ & \trace $\downarrow$ & \diffnorms $\downarrow$ & \cossim $\uparrow$\\
        \toprule
        FedAvg & 52.9 & 297 & 6240 & 11360 & 0.98\\
        FedProx & 53.0 & 270 & 6132 & 6522 & 0.98\\
        MOON & 55.3 & 252 & 5520 & 5712 & 0.97\\
        \midrule
        Mixup & 54.0 & 216 & 5468 & 15434 & \textbf{0.99}\\
        StochDepth & 55.5 & 215 & 3970 & 8267 & 0.97\\
        GradAug ($n=2$) & \textbf{57.1} & 167 & \textbf{2597} & 2924 & 0.96\\
        GradAug ($n=1$) & 56.8 & 179 & 3620 & \textbf{2607} & 0.97\\
        \midrule
        \textbf{\method} & 56.7 & \textbf{143} & 4409 & 9655 & \textbf{0.99} \\
        \bottomrule
    \end{tabular}}
\end{table}

\begin{table}[h]
    \caption{\method~hyperparameter ablations on with CIFAR-100}
    \label{tab:hyper}
    \centering
    \resizebox{\columnwidth}{!}{%
    \begin{tabular}{cccc|ccc}
        \toprule
        Method & $\mu=0.35$ & $\mu=0.45$ & $\mu=0.55$ & $\omega_{S}=0.1$ & $\omega_{S}=0.25$ & $\omega_{S}=0.4$\\
        \toprule
        \method & 56.0 & \textbf{56.7} & 56.1 & 54.9 & \textbf{56.7} & 55.2\\
        \bottomrule
    \end{tabular}
    }
\end{table}

\section{Data Partitioning}
As is common in the literature \cite{moon, feddyn, fedml}, we partition the employed datasets into $K$ unbalanced subsets using a Dirichlet distribution $Dir(\alpha)$. The distribution for all three datasets at $\alpha=0.5$ is visualized in Fig. \ref{fig:data} (a), (c), and (d). Additionally, (b) shows the distribution for CIFAR-100 with $\alpha=0.1$ as studied in Section 3.5 of the main paper. Overall, we see that the number of samples for each class varies considerably across clients, and often times a client will not have any samples from multiple classes. 
This enhances the FL setting by making it more realistic and challenging.
For implementation, we utilize the same data partitioning script as that in \cite{fedml}.

\section{Additional Implementation Details} \label{sec:details}
When calculating $\mathbf{X}_{F}$ and $\mathbf{X}_{S}$, the input and output features involved will typically be of different spatial sizes in practice. Therefore, \cite{london} utilizes an adaptive average pool operation in PyTorch to reduce the spatial size of the larger feature map to that of the smaller one. We likewise employ this operation.

Prior to performing backpropagation, we apply a relative scale to $\mathcal{L}_{Lip}$ along with the $\mu$ scaling parameter. In PyTorch-style pseudocode: $loss\_lip$ $=$ $\mu^{*}$ $(loss\_ce.item()/loss\_lip.item())^{*}$ $loss\_lip$. This is to ensure that $\mathcal{L}_{Lip}$ is on relatively the same scale with $\mathcal{L}_{CE}$.
A gradient clip is also applied.

\end{document}